\documentclass{article}

\PassOptionsToPackage{numbers, compress}{natbib}

\usepackage[final]{neurips_2020}
\usepackage[utf8]{inputenc} 
\usepackage[T1]{fontenc}    
\usepackage{hyperref}      
\usepackage{url}            
\usepackage{booktabs}       
\usepackage{amsfonts}     
\usepackage{nicefrac}       
\usepackage{microtype}     
\usepackage{graphicx}
\usepackage{enumitem}
\usepackage[table]{xcolor}
\newcommand{{\AN}}{TurboViT}
\title{\AN: Generating Fast Vision Transformers via Generative Architecture Search}

\author{Alexander Wong$^{1,2,3,*}$, Saad Abbasi$^{1,2,3,*}$, Saeejith Nair$^{1,2}$ \\
$^1$University of Waterloo, Waterloo, Ontario, Canada\\
$^2$Waterloo Artificial Intelligence Institute, Waterloo, Ontario, Canada\\
$^3$DarwinAI,  Waterloo, Ontario, Canada \\
$^*$equal contributors
}

\begin{document}

\maketitle

\begin{abstract}
Vision transformers have shown unprecedented levels of performance in tackling various visual perception tasks in recent years.  However, the architectural and computational complexity of such network architectures have made them challenging to deploy in real-world applications with high-throughput, low-memory requirements.  As such, there has been significant research recently on the design of efficient vision transformer architectures. In this study, we explore  the generation of fast vision transformer architecture designs via generative architecture search (GAS) to achieve a strong balance between accuracy and architectural and computational efficiency.  Through this generative architecture search process, we create {\AN}, a highly efficient hierarchical vision transformer architecture design that is generated around mask unit attention and Q-pooling design patterns.  The resulting {\AN} architecture design achieves significantly lower architectural computational complexity ($>2.47\times$ smaller than FasterViT-0 while achieving same accuracy) and computational complexity ($>3.4\times$ fewer FLOPs and 0.9\% higher accuracy than MobileViT2-2.0) when compared to 10 other state-of-the-art efficient vision transformer network architecture designs within a similar range of accuracy on the ImageNet-1K dataset. Furthermore, {\AN} demonstrated strong inference latency and throughput in both low-latency and batch processing scenarios (\textbf{$>3.21\times$} lower latency and \textbf{$>3.18\times$} higher throughput compared to FasterViT-0 for low-latency scenario).  These promising results demonstrate the efficacy of leveraging generative architecture search for generating efficient transformer architecture designs for high-throughput scenarios.
\end{abstract}

\section{Introduction}

Vision transformers have shown unprecedented levels of performance in tackling various visual perception tasks in recent years~\cite{vit,deit,deit2,gcvit,swin,swin2,hiera,mvit}.  However, the architectural and computational complexity of such network architectures have made them challenging to deploy in real-world applications with high-throughput, low-memory requirements.  As such, there has been significant research recently on the design of efficient vision transformer architectures~\cite{mobilevit,mobilevit2,fastvit,fastervit,efficientvit,efficientvit2,cvt,pit}.  For example, Cai et al.~\cite{efficientvit} introduced a lightweight multi-scale attention comprised of only lightweight and hardware-efficient operations.  Vasu et al.~\cite{fastvit} introduced a hybrid convolutional-transformer architecture design that leverages structural reparameterization to reduce memory access costs.  Hatamizadeh et al.~\cite{fastervit} introduces a hierarchical attention approach that decomposes global self-attention into multi-level attention to reduce computational complexity.  Another interesting approach is one proposed by Wu et al.~\cite{cvt}, where they not only introduced convolutions to vision transformers, but also employed a data-adapted neural architecture search for discovering efficient vision transformer architecture designs.  It is along this later direction of network architecture search that we will explore for finding efficient vision transformers, but accomplished instead via a generative approach.

In this study, we explore the generation of fast vision transformer architecture designs via generative architecture search (GAS)~\cite{wong2018ferminets,graspnext} to achieve a strong balance between accuracy and architectural and computational efficiency.  Through this generative architecture search process, we create {\AN}, a highly efficient hierarchical vision transformer architecture design that is generated around mask unit attention and Q-pooling design patterns.  The paper is organized as follows.  Section 2 details the generative architecture search strategy used to create {\AN} as well as the resulting generated network architecture design.  Section 3 presents experiments and discussion comparing {\AN} with state-of-the-art efficient vision transformer architecture designs across multiple performance metrics.

\section{Methods}

\textbf{Generative Architecture Search.} In this study, we leveraged generative synthesis~\cite{wong2018ferminets} to conduct generative architecture search (GAS) to identify the architecture design of {\AN}. More specifically, generative synthesis can be formulated as a constrained optimization problem, where the goal is to identify the optimal generator $G^\star(\cdot)$ generating network architectures $N$ that maximizes a universal performance function $U$ (e.g.,~\cite{netscore}) under a given set of operational constraints defined by an indicator function $\textbf{1}_r(\cdot)$, 

\begin{equation}
\mathcal{G}=\max_{\mathcal{G}}\mathcal{U}(\mathcal{G}(s)) \;\;\;\textrm{ subject to} \;\;\; 1_r(G(s))=1, \;\;\forall\in \mathcal{S}.
\end{equation}

\begin{figure}
    \includegraphics[width=\linewidth]{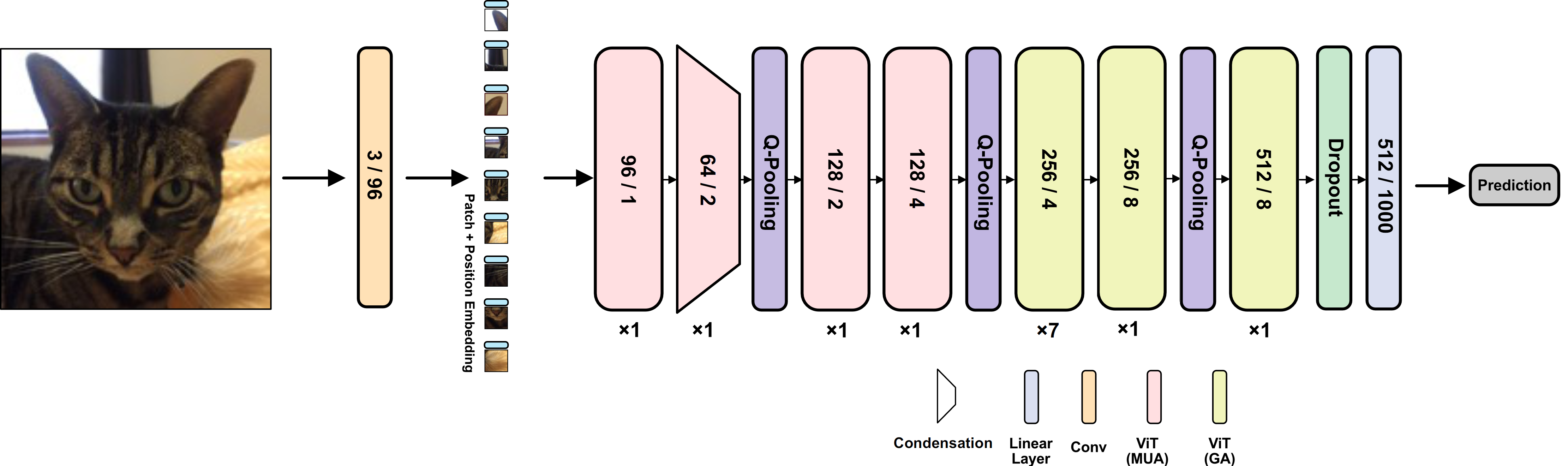}
    \caption{Proposed {\AN} vision transformer architecture design, generated via generative architecture search (GAS) around global attention (GA), mask unit attention (MUA), and Q-pooling design patterns introduced in~\cite{hiera}. The values (D/H) within a ViT block indicates its hidden dimensionality and head count, while numbers below a block indicates the number of sequential blocks of that type. The values (I/O) within a convolution block or a linear layer indicates its input dimensionality and output dimensionality.}
    \label{fig:turbovit_architecture}
\end{figure}

This constrained optimization problem is solved in an iterative manner, with a full detailed description of the process presented in~\cite{wong2018ferminets}. In this study, we impose the following design constraints via $\textbf{1}_r(\cdot)$ to identify a vision transformer architecture design for {\AN} that achieves the desired computational complexity tailored for high-throughput scenarios:
\begin{enumerate}
    \item Leverage both global attention and mask unit attention design patterns as introduced in~\cite{hiera}.  These design patterns have been shown to greatly streamline vision transformer architecture complexities by forgoing vision-specific components without sacrificing accuracy. 
    \item Enforcing the use of Q-pooling design pattern at three locations for reducing architectural and computational complexity via spatial query reduction in a similar way as~\cite{hiera}, thus resulting in a hierarchical architecture design. 
    \item Enforcing a computational complexity constraint of 2.5 GFLOPs to ensure {\AN} has low computational complexity (less than all state-of-the-art vision transformer architecture designs compared in this study) for high-throughput scenarios.
\end{enumerate}

\textbf{Architecture Design.} Figure~\ref{fig:turbovit_architecture} demonstrates the {\AN} architecture design generated via generative architecture search. Overall, it can be observed that the architecture design is quite clean and streamlined when compared to other state-of-the-art efficient vision transformer architecture designs (particularly more complex hybrid convolutional-transformer architecture designs), consisting largely of a sequence of ViT blocks with relatively low hidden dimensionalities as well as relatively low head counts (especially when compared to ViT~\cite{vit}), thus facilitating for greater architectural and computational efficiency.  

As expected, it can be observed that the {\AN} architecture design consists of Q-pooling at three different locations in the architecture design to allow for architectural and computational efficiencies through spatial reductions, with the majority of layers located after the second Q-pooling.  It can also be observed that earlier ViT blocks in  the {\AN} architecture design leverage local attention via mask unit attention while later ViT blocks leverage global attention, thus not harnessing global attention where it is less useful in exchange for significant gains in computational efficiency.

A particularly interesting observation about the {\AN} architecture design is the presence of a hidden dimensionality condensation mechanism imposed at the beginning of the architecture design, where the hidden dimensionality is greatly reduced at the second ViT block to form a highly condensed embedding when compared to the first ViT block before the hidden dimensionality progressively increases as we move down the architecture.  Such a condensation mechanism appears to be effective at greatly reducing computational complexity while still achieving high representational capabilities in the overall architecture design.

\begin{table}[t]
	\centering
	\caption{Top-1 accuracy, number of parameters, and number of FLOPs of {\AN} in comparison to 10 state-of-the-art efficient vision  transformer architecture designs with a similar range of accuracy for image classification on the  ImageNet-1K dataset.  Best results are in \textbf{bold}.}
	\begin{tabular}{p{4cm}cccc}
		\hline
		\textbf{Model}& \textbf{Top-1 Accuracy} & \textbf{\# Parameters} &	\textbf{FLOPs} \\
		\textbf{}& \textbf{(\%)} & \textbf{(M)} &	\textbf{(G)} \\  
		\hline 
		MobileViTv2-2.0~\cite{mobilevit2}		& 	81.2
	&	18.5	&	7.5 \\
  		MViTv2-T~\cite{mvit}		& 	82.3	&	24.0	&	5.0 \\
		Swin-T~\cite{swin}		& 	81.3	&	29.0	&	4.5 \\
		SwinV2-T~\cite{swin2}		& 	81.8	&	28.3	&	4.4 \\
		CvT-13-NAS~\cite{cvt}		& 	82.2	&	18.0	&	4.1 \\
		FastViT-SA24~\cite{fastvit}		& 	\textbf{82.6}	&	20.6	&	3.8 \\
		LITv2-S~\cite{LITv2}		& 	82.0	&	28.0	&	3.7 \\  
		FasterViT-0~\cite{fastervit}		& 	82.1	&	31.4	&	3.3 \\
		PiT-S~\cite{pit}		& 	81.9	&	23.5	&	2.9 \\    
		Twins-S~\cite{twins}		& 	81.7	&	24.1	&	2.8 \\  
		\cellcolor{gray!10}\textbf{\AN}		& 	\cellcolor{gray!10}82.1	&	\cellcolor{gray!10}\textbf{12.7}	&	\cellcolor{gray!10}\textbf{2.2} \\  
		\hline
	\end{tabular}\\
	\label{tab_Results}
\end{table}	

\section{Results and Discussion}
 
The efficacy of the proposed {\AN} architecture design is evaluated on ImageNet-1K dataset and is compared with 10 different state-of-the-art efficient vision transformer architecture designs within a similar range of accuracy for image classification (MobileViTv2-2.0~\cite{mobilevit2}, MViTv2-T~\cite{mvit}, Swin-T~\cite{swin}, SwinV2-T~\cite{swin2}, CvT-13-NAS~\cite{cvt}, FastViT-SA24~\cite{fastvit}, LITv2-S~\cite{LITv2}, FasterViT-0~\cite{fastervit}, PiT-S~\cite{pit}, and Twins-S~\cite{twins}) across three metrics: 1) Top-1 accuracy, 2) architectural complexity (based on number of parameters), and 3) computational complexity (based on number of FLOPs). Furthermore, for {\AN}, FastViT-SA24~\cite{fastvit}, and FasterViT-0~\cite{fastervit}, we also compare inference latency and throughput on an Nvidia RTX A6000 GPU across two difference scenarios.

\textbf{Architectural Complexity.} Table~\ref{tab_Results} shows a comparison between the proposed {\AN} architecture design with the 10 different state-of-the-art efficient vision transformer architecture designs.  In terms of architectural complexity, {\AN} is significantly smaller than all other state-of-the-art efficient vision transformer architecture designs in this comparison.  For example, {\AN} is \textbf{>2.47$\times$} smaller than FasterViT-0 while achieving same accuracy, and \textbf{>1.45$\times$} smaller than MobileViT2-2.0 while at the same time achieving \textbf{0.9\%} higher accuracy. Even when compared against CvT-13-NAS, the second smallest vision transformer architecture design and also the only other design that is created using network architecture search in this study, {\AN} is \textbf{>1.41$\times$} smaller while achieving similar accuracy.  

\textbf{Computational Complexity.} In terms of computational complexity, {\AN} requires significantly fewer FLOPs than all other state-of-the-art efficient vision transformer architecture designs in this comparison.  For example, {\AN} requires \textbf{>3.4$\times$} fewer FLOPs than MobileViT2-2.0 while at the same time achieving \textbf{0.9\%} higher accuracy. When compared to the only other design that is created using network architecture search in this study (CvT-13-NAS), {\AN}  requires \textbf{>1.86$\times$} fewer FLOPs. Even when compared against Twins-S, the vision transformer architecture design requiring the second fewest FLOPs, {\AN} requires \textbf{>1.27$\times$} fewer FLOPS and is \textbf{>1.89$\times$} smaller while achieving \textbf{0.4\%} higher accuracy.  

\textbf{Accuracy.} In terms of accuracy, {\AN} achieves strong top-1 accuracy at significantly lower architectural and computational complexity than all other state-of-the-art efficient vision transformer architecture designs in this comparison.  For example, {\AN} achieves \textbf{0.8\%} and \textbf{0.3\%} higher accuracy when compared to Swin-T and SwinV2-T, respectively, even though it requires \textbf{>2.04$\times$} and \textbf{2$\times$} fewer FLOPs, respectively.  When compared to the only other design that is created using network architecture search in this study (CvT-13-NAS), {\AN} achieves very similar top-1 accuracy (only 0.1\% lower) but at much lower architectural and computational complexity (\textbf{>$1.41\times$} smaller and \textbf{$>1.86\times$} fewer FLOPs, respectively).  Finally, {\AN} achieves a top-1 accuracy that is only 0.5\% lower than the top-performing state-of-the-art efficient vision transformer architecture design in this study (FastViT-SA24).  However, it is important to note that {\AN} is \textbf{>$1.62\times$} smaller and requires \textbf{>$1.72\times$} fewer FLOPs than FastViT-SA24.

 \begin{table}[t]
	\centering
	\caption{Results for inference latency and throughput of {\AN} for two different scenarios (low-latency processing (batch size of 1) and batch processing (batch size of 32) for 224$\times$224 sized RGB image input in comparison to 2 state-of-the-art efficient vision transformer architecture designs (FastViT-SA24~\cite{fastvit} and FasterViT-0~\cite{fastervit}). All evaluations were conducted on an Nvidia RTX A6000 GPU for $N=1000$ runs. Best results are in \textbf{bold}.}
	\begin{tabular}{p{4cm}cccc}
		\hline
		\textbf{Model}& \textbf{Batch Size} & \textbf{Latency} &	\textbf{Throughput} \\
		\textbf{}& \textbf{} & \textbf{(ms)} &	\textbf{(images/s)} \\  
		\hline 
		FastViT-SA24~\cite{fastvit}		& 	1 & 6.3	&	157.9 \\
  		FasterViT-0~\cite{fastervit}		& 	1	&	12.2	&	75.8 \\
		\cellcolor{gray!10}\textbf{\AN}		& 	\cellcolor{gray!10}1	&	\cellcolor{gray!10}\textbf{3.8}	&	\cellcolor{gray!10}\textbf{241.3} \\  
\hline
		FastViT-SA24~\cite{fastvit}		& 	32 & 28.9	&	1089.6 \\
  		FasterViT-0~\cite{fastervit}		& 	32	&	19.0	&	1663.5 \\
		\cellcolor{gray!10}\textbf{\AN}		& 	\cellcolor{gray!10}32	&	\cellcolor{gray!10}\textbf{18.8}	&	\cellcolor{gray!10}\textbf{1696.9} \\  
		\hline
	\end{tabular}\\
	\label{tab_latency}
\end{table}	

\textbf{Inference Latency and Throughput.}  In terms of inference latency and throughput, we evaluated two different scenarios: 1) low-latency processing (using batch size of 1), and 2) batch processing (using batch size of 32).  Table~\ref{tab_latency} shows a comparison for inference latency and throughput on an Nvidia RTX A6000 GPU between {\AN}, FasterViT-0~\cite{fastervit}, and FastViT-SA24~\cite{fastvit}.  It can be observed that, for the low-latency processing scenario, {\AN} significantly outperforms both FasterViT-0~\cite{fastervit} and FastViT-SA24~\cite{fastvit}, with latency being \textbf{>3.21$\times$} and \textbf{>1.66$\times$} better, respectively. Furthermore, the throughput of {\AN} for this scenario is \textbf{>3.18$\times$} and \textbf{>1.53$\times$} higher than FasterViT-0~\cite{fastervit} and FastViT-SA24~\cite{fastvit}, respectively.  It can be observed that, for the batch processing scenario, {\AN} significantly outperforms FastViT-SA24, with latency being \textbf{>1.54$\times$} lower and throughput being \textbf{>1.56$\times$} higher.  Interestingly, both {\AN} and FasterViT-0 achieved comparable latency and throughput in this scenario.

These results demonstrate that {\AN} strikes a good balance between between accuracy, architectural complexity, computational complexity as well as demonstrating strong latency and throughput for both low-latency and batch processing scenarios, making it well-suited for high-throughput use cases.  Furthermore, these promising results demonstrate the efficacy of leveraging generative architecture search for generating efficient transformer architecture designs for high-throughput scenarios.

{\small
\bibliographystyle{unsrtnat}
\bibliography{main}
}

\end{document}